\documentclass{article}
\usepackage{spconf,amsmath,epsfig,bm,graphicx,multirow,subfigure,microtype,hyperref}
\begin{document}\sloppy

\def\x{{\mathbf x}}
\def\L{{\cal L}}

\title{Quality-Gated Convolutional LSTM for Enhancing Compressed Video}
%
\name{Ren Yang*$^3$, Xiaoyan Sun$^1$, Mai Xu$^2$ and Wenjun Zeng$^1$ \thanks{*This work was done when Ren Yang was an intern in Microsoft Research. Corresponding authors: Xiaoyan Sun and Mai Xu. This work is partly supported by NSFC under Grants 61876013 and 61573037, and Fok Ying Tung Education Foundation under Grant 151061.}}
\address{$^1$Microsoft Research \quad $^2$Hangzhou Innovation Institute, Beihang University \quad $^3$ETH Z$\ddot{\text{u}}$rich \\ r.yangchn@gmail.com, xysun@microsoft.com, maixu@buaa.edu.cn, wezeng@microsoft.com}

\maketitle

\begin{abstract}
The past decade has witnessed great success in applying deep learning to enhance the quality of compressed video. However, the existing approaches aim at quality enhancement on a single frame, or only using fixed neighboring frames. Thus they fail to take full advantage of the inter-frame correlation in the video. This paper proposes the Quality-Gated Convolutional Long Short-Term Memory (QG-ConvLSTM) network with bi-directional recurrent structure to fully exploit the advantageous information in a large range of frames. More importantly, due to the obvious quality fluctuation among compressed frames, higher quality frames can provide more useful information for other frames to enhance quality. Therefore, we propose learning the ``forget'' and ``input'' gates in the ConvLSTM cell from quality-related features. As such, the frames with various quality contribute to the memory in ConvLSTM with different importance, making the information of each frame reasonably and adequately used. Finally, the experiments validate the effectiveness of our QG-ConvLSTM approach in advancing the state-of-the-art quality enhancement of compressed video, and the ablation study shows that our QG-ConvLSTM approach is learnt to make a trade-off between quality and correlation when leveraging multi-frame information. The project page: \url{https://github.com/ryangchn/QG-ConvLSTM.git}
\end{abstract}
\vspace{-.5em}
\section{Introduction}\label{intro}
\vspace{-.5em}
Nowadays, video becomes more and more popular in multimedia applications.
When transmitting video over the bandwidth-limited channel, video compression has to be applied to significantly save the bit-rate. However, the compressed video inevitably incurs compression artifacts, which may lead to severe degradation on the Quality of Experience (QoE). Therefore, it is necessary to study on enhancing the visual quality of compressed video.

During the past decade, an increasing number of works have focused on the quality enhancement of compressed image and video. Most of them \cite{foi2007pointwise,jung2012image,chang2014reducing,dong2015compression,Guo2016Building, wang2016d3,Zhang2017Beyond, li2017efficient,Tai2017MemNet} aim at improving the visual quality of JPEG image. Specifically, \cite{foi2007pointwise,jung2012image,chang2014reducing} utilized non-deep-learning methods for JPEG restoration.
Then, Dong \textit{et al.} \cite{dong2015compression} pioneered on applying deep networks for enhancing JPEG image quality
Afterwards, $\text{\textbf{D}}^3$ \cite{wang2016d3} and DDCN \cite{Guo2016Building} were proposed for removing the JPEG artifacts, taking the prior knowledge of JPEG into consideration. Later, DnCNN was proposed in \cite{Zhang2017Beyond} for several image restoration tasks. Most recently, the MemNet \cite{Tai2017MemNet}, which adopts the memory blocks, becomes the state-of-the-art image enhancement method.

\begin{figure}[!t]
\centering
\vspace{-1em}
\includegraphics[width=.95\linewidth]{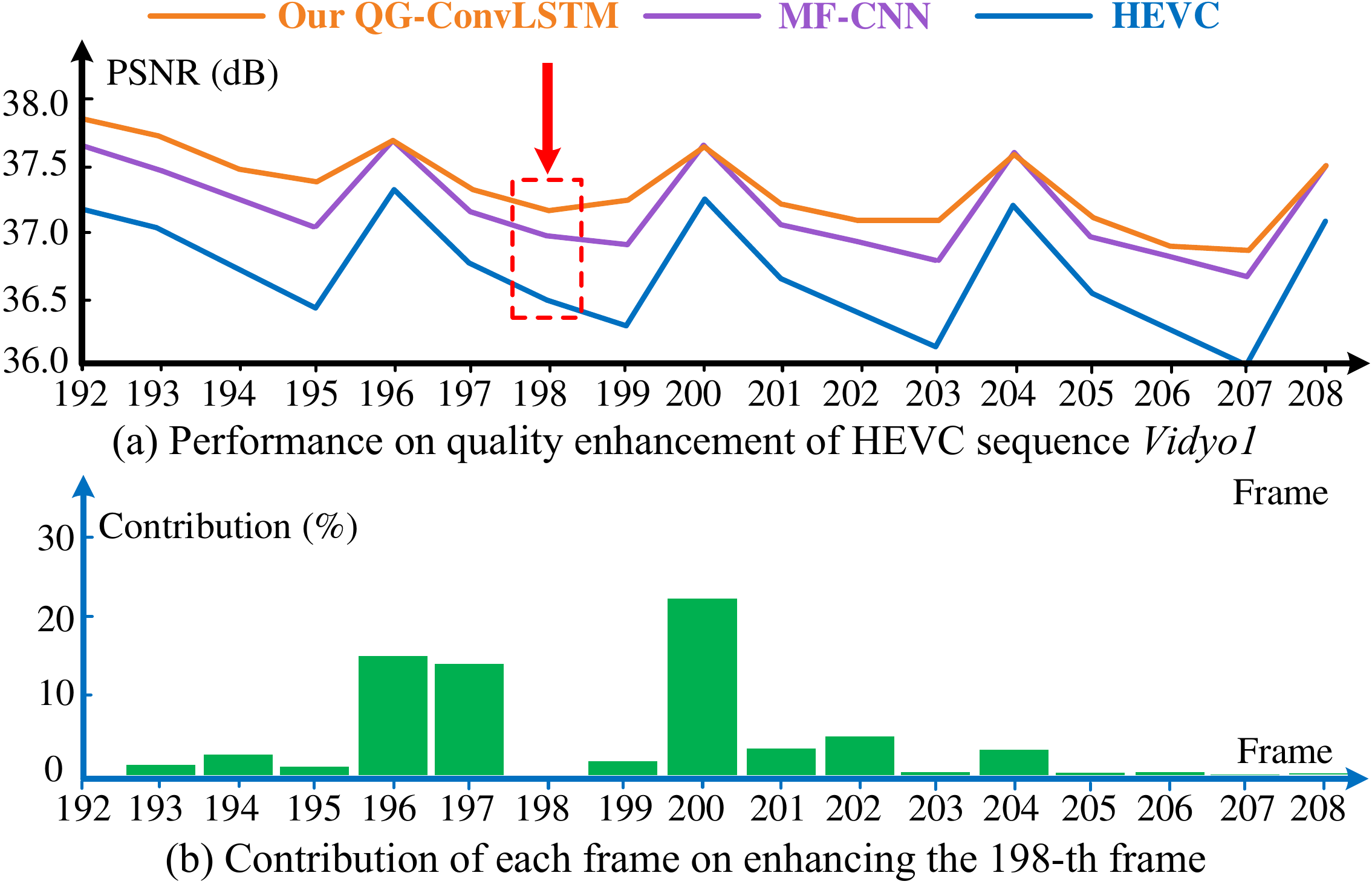}
\vspace{-1em}
\caption{Example result of our QG-ConvLSTM approach}\label{1}
\vspace{-1em}
\end{figure}

For enhancing the quality of compressed video, the VRCNN \cite{dai2017convolutional} was designed as a post-processing filter for HEVC intra-coding. Later, Wang \textit{et al.} \cite{Wang2017A} proposed enhancing the decoded HEVC video via a CNN-based method, called DCAD. 
Recently, the QE-CNN method \cite{yang2017decoder,yang2017enhancing} was proposed, in which the QE-CNN-I and QE-CNN-P networks are learnt to handle the intra- and inter-coding distortion, respectively. However, all of the above are single-frame methods, i.e., only one frame is input to the network once. Therefore, these methods fail to take the inter-frame correlation into consideration, which severely limits their performance. To overcome this shortcoming, Yang \textit{et al.} proposed the Multi-Frame CNN (MF-CNN) for video enhancement.
In MF-CNN, Peak Quality Frames (PQFs)\footnote{PQF is defined as the frame with higher quality than its previous and subsequent frames, e.g., frames 196, 200 and 204 in Fig. \ref{1}-(a).} are applied to help the neighboring non-PQFs to enhance quality. 
Nevertheless, the MF-CNN only makes use of a finite number of fixed frames, i.e., the two nearest PQFs. Hence, it ignores the useful information available in other frames, and does not take the trade-off between quality and correlation into consideration.

In this paper, the Quality-Gated Convolutional Long Short-Term Memory (QG-ConvLSTM) network is proposed for enhancing the quality of compressed video. In our QG-ConvLSTM approach, the ConvLSTM \cite{xingjian2015convolutional} structure is utilized to fully exploit the advantageous information in the whole video. Moreover, because of the quality fluctuation among compressed frames, frames with different quality should be unequally important for helping other frames to enhance quality. Therefore, we propose learning the weights of the ``forget'' and ``input'' gates in ConvLSTM from quality-related features by another LSTM network, replacing those gates in the original ConvLSTM cell. As such, the weights to forget the previous memory and to update the current information to the memory cell are guided by the compressed quality. For example, high quality frames are expected to forget the lower quality information in previous memory and update its high quality information to the memory to assist the enhancement of other frames. Consequently, our QG-ConvLSTM approach learns to reasonably and adequately leveraging multi-frame information for quality enhancement of compressed video. Note that since raw videos are unavailable at decoder side, in this paper, we use the no-reference quality-related features, which are extracted from compressed frames and video decoder.

An example of our QG-ConvLSTM approach is illustrated in Fig. \ref{1}. As shown in Fig. \ref{1}, a large number of frames contribute to enhancing frame 198 in our QG-ConvLSTM approach. Among them, frames 196, 197 and 200 are with higher quality than frame 198 and also with low temporal distance, so they contribute most on enhancing frame 198. Besides, compared with frame 201, frame 204 is less correlated with frame 198, but its quality is obviously higher than frame 201. As a result, frames 201 and 204 have almost the same contribution to frame 198. Note that the contribution is calculated by the formula introduced in Section \ref{abl}. Finally, we can see from Fig. \ref{1}-(a) that compared with the MF-CNN, our QG-ConvLSTM approach achieves better performance and lower fluctuation of enhanced quality. The main contributions of this paper are:

(1) We propose a ConvLSTM network for quality enhancement of compressed video, fully taking advantage of the useful information available in the video.

(2) We propose gating the ConvLSTM network by learnt weights from quality-related features, to reasonably make use of the information in frames with various compressed quality.

\begin{figure}[!t]
\centering
\includegraphics[width=.85\linewidth]{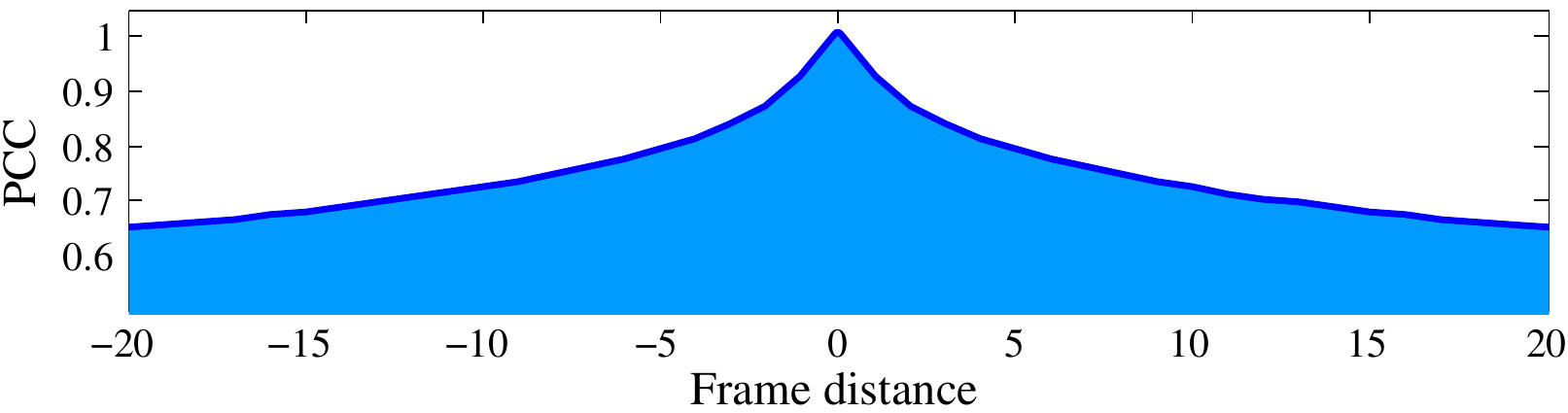}
\vspace{-1em}
\caption{Average PCC between frames with various distances}\label{pcc}
\vspace{-1em}
\end{figure}
\vspace{-.5em}
\section{Preliminary}\label{pre}
\vspace{-.5em}
Since our QG-ConvLSTM approach is motivated by the high correlation among continuous frames and the obvious variation of compressed quality, in this section, we analyze the correlation between video frames and the quality fluctuation of compressed video alongside the frames. In this paper, we follow \cite{Yang2018Multi} to use the database of 70 videos (denoted as Vid-70), which are encoded by various standards, including MPEG-1, MPEG-2, MPEG-4, H.264/AVC and HEVC.
\vspace{-.5em}
\subsection{Content correlation}\label{cc}
\vspace{-.5em}
The correlation between two frames is evaluated in terms of Pearson Correlation Coefficient (PCC). Here, we calculate the PCC between each frame and its 40 neighboring (20 previous and 20 subsequent) frames. Fig. \ref{pcc} shows the PCC values averaged among all frames in Vid-70 database. It can be seen from Fig. \ref{pcc} that the PCC values are larger than 0.79 within 5 frames. At the interval of 10 frames, the averaged PCC between is also higher than 0.72. Such figure is about 0.65 when the distance enlarges to 20 frames. These verify that the frames in a large range exist strong correlation in contents, and such correlation degrades along frames interval.

Accordingly, fully exploiting the information in a large range of frames
may obviously improve the performance on enhancing compressed video. Hence, we propose adopting the bi-directional ConvLSTM structure in this paper.

\vspace{-.5em}
\subsection{Quality fluctuation}\label{qf}
\begin{figure}[!t]
\centering
\includegraphics[width=1\linewidth]{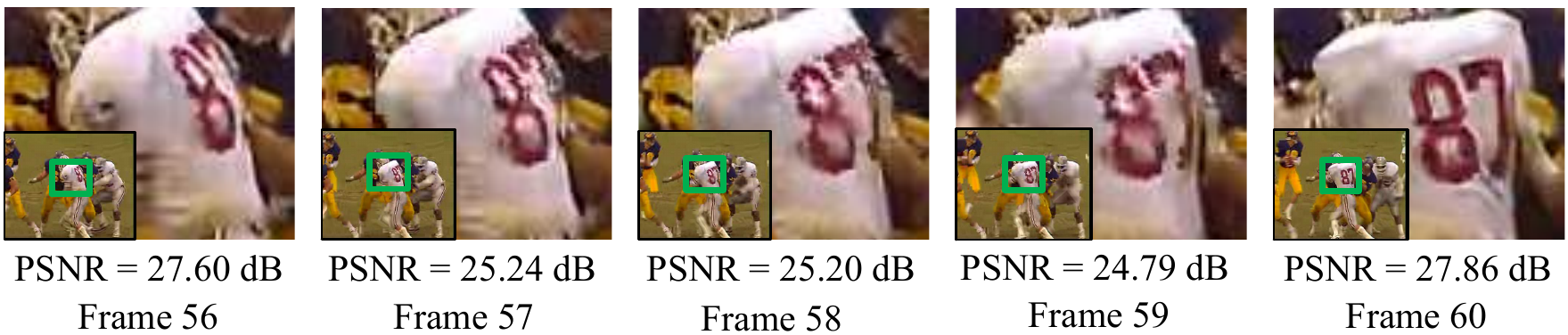}
\vspace{-2em}
\caption{Quality fluctuation in HEVC video \textit{Football} \cite{Yang2018Multi}}\label{fluc}
\vspace{-1.5em}
\end{figure}

\vspace{-.5em}
According to the analysis in \cite{Yang2018Multi}, the frame quality remarkably fluctuates after compression. The compressed quality is evaluated via Peak Signal-to-Noise Ratio (PSNR). Specifically, in the Vid-70 database, the averaged STandard Deviation (STD) of frame-level PSNR for each compressed video is 1.83 dB for MPEG-1/2, 1.78 dB for MPEG-4, 1.64 dB for H.264 and 1.06 dB for HEVC, respectively.
The Peak-Valley Differences (PVD), which indicates the differences between the nearest peak and valley values in PSNR curves, is also higher than 1.00 dB for MPEG-1/2/4 and H.264. The averaged PVD is as large as 1.51 dB for the latest HEVC standard. This validates the large quality difference alongside compressed frames. Fig. \ref{fluc} illustrates an example, showing the obvious quality fluctuation among compressed frames.

\begin{figure*}[!t]
\vspace{-2em}
\centering
  \subfigure[Overall framework]{\includegraphics[width=.68\linewidth]{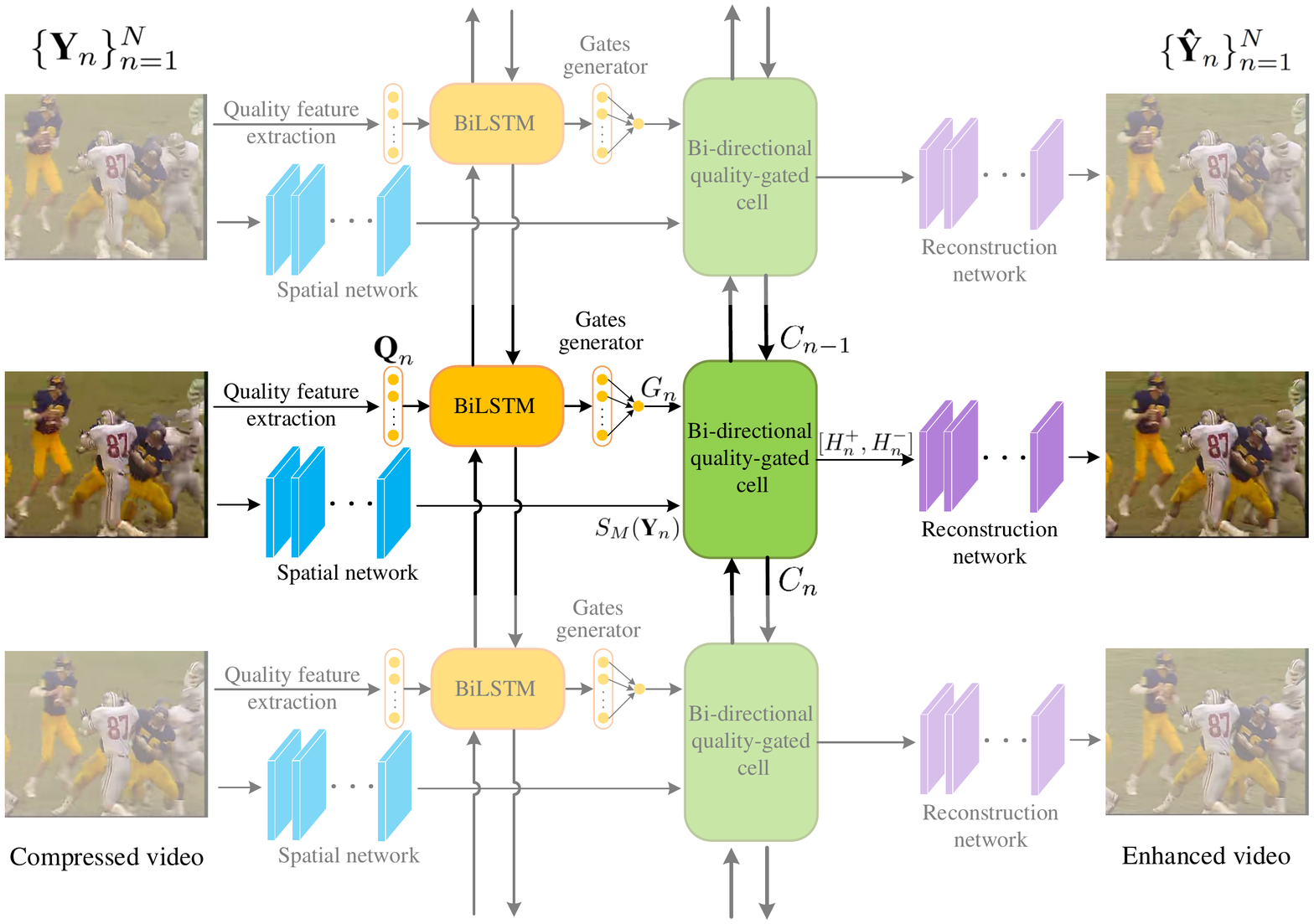}}
  \hspace{1em}
  \subfigure[Structure of the quality-gated cell]{\includegraphics[width=.28\linewidth]{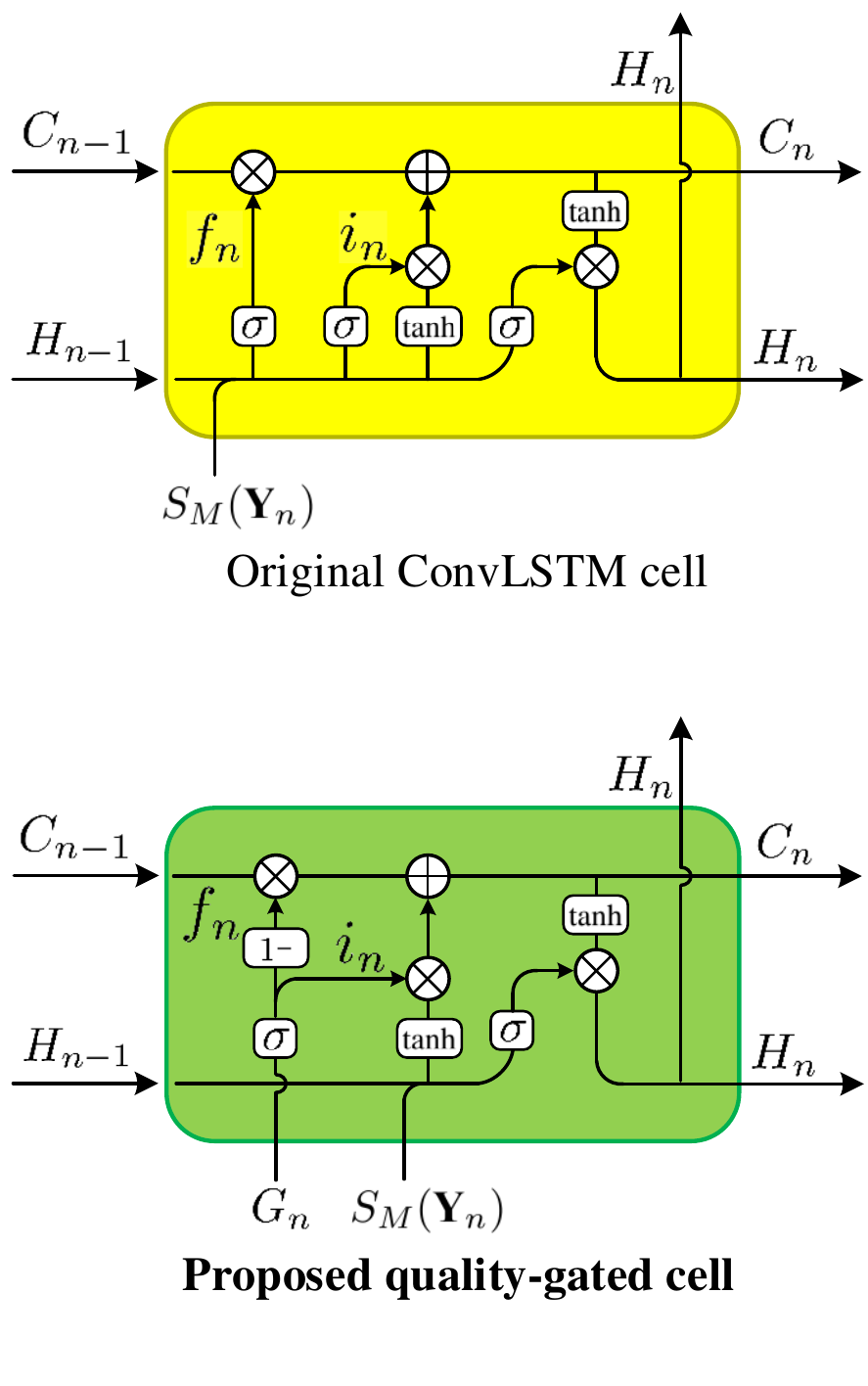}}
  \vspace{-1em}
  \caption{The framework of our QG-ConvLSTM approach and the proposed quality-gated cell}\label{framework}
  \vspace{-1em}
\end{figure*}

As a result, when utilizing the ConvLSTM to enhance compressed quality, the quality fluctuation should also be considered. That is, the frames with various quality should contribute differently on providing useful information to other frames.
Motivated by this, we propose gating the ConvLSTM by the weights learnt from quality-related features.

\vspace{-.5em}
\section{Proposed QG-ConvLSTM approach}\label{app}
\vspace{-.5em}
\subsection{Framework}\label{fw}
\vspace{-.5em}
Fig. \ref{framework} illustrates the framework of our QG-ConvLSTM approach. Overall speaking, our QG-ConvLSTM network is designed as a spatial-temporal structure, which adopts the bi-directional ConvLSTM structure to fully exploit the information in both previous and incoming frames. More importantly, we propose generating the weights of the ``forget'' ($f_n$) and ``input'' ($i_n$) gates in the ConvLSTM cell by an 1D LSTM network, with the input of quality-related features. Therefore, the compressed quality guides the ratios for forgetting the previous memory and updating the current information in the ConvLSTM. As a result, the frames with various compressed quality contributes to the memory cell in ConvLSTM with different significance. This makes the information with different quality reasonably used. 

Specifically, our QG-ConvLSTM approach contains four components, i.e., the spatial network, the gates generator, the quality-gated cell and the reconstruction network. The functions of these four networks are denoted as $F_S(\cdot)$, $F_G(\cdot)$, $F_C(\cdot)$ and $F_R(\cdot)$, respectively. Besides, we define the compressed video as $\{\textbf{Y}_n\}^{N}_{n=1}$, where $n$ indicates the frame index and $N$ is the total number of frames. As such, defining the quality features as $\textbf{Q}_n$ at the time step of $n$, the function of our QG-ConvLSTM approach can be expressed as
\begin{eqnarray}
\{\textbf{\^{Y}}_n\}^{N}_{n=1} = F_R(F_C(F_S(\{\textbf{Y}_n\}^{N}_{n=1}), F_G(\{\textbf{Q}_n\}^{N}_{n=1}))),
\end{eqnarray}
where $\{\textbf{\^{Y}}_n\}^{N}_{n=1}$ indicates the output frames with enhanced quality. In following, we introduce each part of our QG-ConvLSTM approach.

\textbf{Spatial network.} As Fig. \ref{framework}-(a) shows, we adopt CNN layers with the activation function of Rectified Linear Unit (ReLU) \cite{he2015delving} in the spatial network, to extract spatial features from the compressed frames. Specifically, let $W_{sm}$ and $B_{sm}$ denote the weights and bias matrices of the $m$-th convolutional layer, the expression of our spatial network for the $n$-th compressed frame $\textbf{Y}_n$ is as follows
\begin{align}
&S_0(\textbf{Y}_n)=\textbf{Y}_n,\\
&S_m(\textbf{Y}_n)=\text{ReLU}(W_{sm}\ast S_{m-1}(\textbf{Y}_n) + B_{sm}), 1\leq m\leq M, \label{cnn}
\end{align}
in which the total number of CNN layers is denoted as $M$. Consequently, as a temporal sequence,
\begin{eqnarray}
F_S(\{\textbf{Y}_n\}^{N}_{n=1}) = \{S_M(\textbf{Y}_n)\}^{N}_{n=1}
\end{eqnarray}
exists. In our spatial network, the CNN layers for each time step $n$ share the parameters of $\{W_{sm}\}^M_{m=1}$ and $\{B_{sm}\}^{M}_{m=1}$.

\textbf{Gates generator.} Recall that the raw video cannot be obtained in quality enhancement, the no-reference features are used in our gates generator. In this paper, we follow \cite{Yang2018Multi} to utilize the 36 spatial features extracted by the no-reference quality assessment method \cite{mittal2012no}. Besides, the Quantization Parameter (QP) and the bit allocation are also applied in our approach as compression domain features, which can be obtained directly from the video decoder. As such, for the $n$-th frame, we get a 38-dimensional quality-related feature, denoted as $\bm{q}_n$. The significance of a frame for enhancing other frames is decided by its relative quality compared with others, rather than the absolute quality. Hence, for the $n$-th frame, we input the quality features of the current and $T$ neighboring frames to our gates generator network. That is, we have
\begin{align}
\textbf{Q}_n = \{\bm{q}_{n-T/2},...,\bm{q}_n,...,\bm{q}_{n+T/2}\} \label{qn}
\end{align}
as the input feature with $38\cdot(T+1)$ dimensions. 

Fig. \ref{framework}-(a) shows that, in our gates generator, the Bi-directional LSTM (BiLSTM), which is able to learn the temporal feature of quality fluctuation, is applied. Then, the output of the forward and backward LSTM networks are concatenated and input to a fully-connected layer to predict the gate weights for our quality-gated cell. To summarize, let ``+'' and ``-'' indicate the directions of forward and backward, the expression of our gates generator can be written as
\begin{eqnarray}
\{h^{+}_n\}^{N}_{n=1} &=& \text{LSTM}^{+}(\{\textbf{Q}_n\}^{N}_{n=1}),\\
\{h^{-}_n\}^{N}_{n=1} &=& \text{LSTM}^{-}(\{\textbf{Q}_n\}^{N}_{n=1}),\\
G_n &=& W_{fc}\cdot[h^{+}_n, h^{-}_n] + B_{fc}\label{fc}
\end{eqnarray}
where $h_n$ is defined as the output of the LSTM network at the time step of $n$, and $[\cdot,\cdot]$ represents the concatenation along channels. In \eqref{fc}, $W_{fc}$ and $B_{fc}$ are the weight and bias matrices of the fully-connected layer, and $G_n$ is the output of the $n$-th frame. The same as \eqref{cnn}, $W_{fc}$ and $B_{fc}$ share parameters for each time step. Finally, we obtain $F_G(\{\textbf{Q}_n\}^{N}_{n=1}) = \{G_n\}^{N}_{n=1}$ as the learnt gate weights for the quality-gated cell in our QG-ConvLSTM approach. The quality-gated cell will be introduced in the following. 

\textbf{Quality-Gated cell.} As shown in Fig. \ref{framework}-(b), in original ConvLSTM cell, with the input of $S_M(\textbf{Y}_n)$ in \eqref{cnn} and the output of the last step (denoted as $H_{n-1}$), the memory cell at the $n$-th time step is formulated as
\begin{align}
C_n = f_n\cdot C_{n-1} + i_n\cdot\text{tanh} (W_{c}\ast [S_M(\textbf{Y}_n), H_{n-1}] + B_{c}), \label{mem}
\end{align}
where $W_c$ and $B_c$ are the weights and bias of the convolutional layer, respectively. In \eqref{mem}, $f_n$ indicates the weight of the ``forget'' gate, and $i_n$ is for the ``input'' gate. As we can see from \eqref{mem}, small $f_n$ makes $C_n$ forget the former memory, and large $i_n$ updates the current information.

As mentioned above, in our QG-ConvLSTM approach, the importance of each frame is distinguished via the compressed quality, because higher quality frames may contain more useful information to help to enhance other frames. Thus, we apply the output of \eqref{fc}, i.e., $G_n$, to gate $C_n$ in our quality-gated cell, replacing the original ``forget'' and ``input'' gates. Assume that the $n$-th frame is with high quality, then we intend to let $C_n$ to forget the previous information (with small $f_n$), since the former information is with lower quality and also less correlated with the upcoming frames. Meanwhile, the information of this high quality frame should be updated to $C_n$ (with large $i_n$) to provide advantageous information for subsequent frames. On the contrary, when a low quality frame comes, the network is expected to not forget the previous memory (i.e., large $f_n$) and update little current information (i.e., small $i_n$). Accordingly, we set
\begin{eqnarray}
f_n = 1 - \sigma(G_n)\quad\text{and}\quad i_n = \sigma(G_n),
\end{eqnarray}
in our quality-gated cell, where $\sigma$ is the sigmoid function to constraint $f_n$ and $i_n$ to the range of $(0,1)$. As a result, the gates to forget and update the memory are controlled by $G_n$, which is learnt from the features of compressed quality.

In conclusion, the proposed quality-gated cell can be expressed as
\begin{eqnarray}
\tilde{C}_n &=& \text{tanh} (W_{c}\ast [S_M(\textbf{Y}_n), H_{n-1}] + B_{c}),\\
C_n &=& (1 - G_n)\cdot C_{n-1} + G_n\cdot\tilde{C}_n,\\
O_n &=& \sigma(W_{o}\ast [S_M(\textbf{Y}_n), H_{n-1}] + B_{o}),\\
H_n &=& O_n\circ \text{tanh}(C_n),
\end{eqnarray}
where $\circ$ denotes the Hadamard products, and $H_n$ is the output of our quality-gated cell. Recall that we apply the bi-directional ConvLSTM structure in our
QG-ConvLSTM approach, let $H^{+}_n$ and $H^{-}_n$ denote the forward and backward outputs of our quality-gated cell. Then,
\begin{eqnarray}
F_C(F_S(\{\textbf{Y}_n\}^{N}_{n=1}), F_G(\{\textbf{Q}_n\}^{N}_{n=1})) = \{[H^{+}_n, H^{-}_n]\}^{N}_{n-1}
\end{eqnarray}
can be obtained.

\textbf{Reconstruction network.} Finally, we adopt an $L$-layer CNN to reconstruct the enhanced frames of compressed video. That is,
\begin{align}
&R_0(H_n) = [H^{+}_n, H^{-}_n], \\
&R_l(H_n) = \text{ReLU}(W_{rl}\ast R_{l-1}(H_n) + B_{rl}),\quad 1\leq l\leq L-1, \\
&R_L(H_n) = W_{rL}\ast R_{L-1}(H_n) + B_{rL},
\end{align}
where $W_{rl}$ and $B_{rl}$ are the weights and biases for the reconstruction CNN layers. Similar to the spatial feature extractor, the parameters of $\{W_{rl}\}^{L}_{l=1}$ and $\{B_{rl}\}^{L}_{l=1}$ are shared for all time steps. Consequently, we get the enhanced video as
\begin{eqnarray}
\{\textbf{\^{Y}}_n\}^{N}_{n=1} = \{R_L(H_n)\}^{N}_{n=1}.
\end{eqnarray}

\begin{table*}[!t]
  \centering
  \vspace{-3em}
  \footnotesize
  \caption{Quality enhancement performance in terms of $\Delta$PSNR (dB)}
    \begin{tabular}{|c|c|c|c|c|c|c|c|c|c|}
    \hline
    \multirow{2}[2]{*}{QP} & \multirow{2}[2]{*}{Resolution} & \multirow{2}[2]{*}{Seq.} & AR-CNN \cite{dong2015compression}& Li \textit{et al.} \cite{li2017efficient} & DnCNN \cite{Zhang2017Beyond} & DCAD \cite{Wang2017A}& QE-CNN \cite{yang2017enhancing}& MF-CNN \cite{Yang2018Multi}& Proposed \\
         & \multicolumn{1}{c|}{} &       & (ICCV'15) & (ICME'17) & (TIP'17) & (DCC'17) & (TCSVT'18) &(CVPR'18)& QG-ConvLSTM \\
    \hline
    \multirow{11}[2]{*}{42} & $352\times 288$  & 1 & 0.2275 & 0.2242 & 0.2523 & 0.2140 & 0.2664 & 0.3968 & \textbf{0.5686} \\
\cline{2-10}          & $416\times 240$  & 2 & 0.1509 & 0.1876 & 0.1970 & 0.1556 & 0.1986 & 0.4506 & \textbf{0.6066} \\
\cline{2-10}          & $416\times 240$  & 3 & 0.1899 & 0.2266 & 0.2301 & 0.1954 & 0.2180 & 0.3908 & \textbf{0.4062} \\
\cline{2-10}          & $1280\times 720$  & 4 & 0.2334 & 0.2789 & 0.2999 & 0.2262 & 0.4161 & 0.4647 & \textbf{0.6352} \\
\cline{2-10}          & $1280\times 720$  & 5 & 0.1457 & 0.2018 & 0.2013 & 0.1358 & 0.4169 & 0.4225 & \textbf{0.5824} \\
\cline{2-10}          & $1280\times 720$  & 6 & 0.1715 & 0.2094 & 0.2233 & 0.0444 & 0.3370 & 0.3359 & \textbf{0.4750} \\
\cline{2-10}          & $1920\times 1080$ & 7 & 0.1073 & 0.1024 & 0.1391 & 0.1164 & 0.3032 & 0.4609 & \textbf{0.5468} \\
\cline{2-10}          & $1920\times 1080$ & 8 & 0.0766 & 0.1550 & 0.1211 & 0.0032 & 0.1396 & 0.5378 & \textbf{0.7412} \\
\cline{2-10}          & $1920\times 1080$ & 9 & 0.1864 & 0.1817 & 0.2001 & 0.0589 & 0.2129 & 0.4223 & \textbf{0.5371} \\
\cline{2-10}          & $2560\times 1600$ & 10 & 0.1382 & 0.1619 & 0.2092 & 0.1324 & 0.5609 & 0.7280 & \textbf{0.9077} \\
\cline{2-10}          & \multicolumn{2}{c|}{\textbf{AVERAGE}} & 0.1627 & 0.1930 & 0.2073 & 0.1282 & 0.3070 & 0.4610 & \textbf{0.6007} \\
    \hline
    \hline
              37      & \multicolumn{2}{c|}{\textbf{AVERAGE}} & 0.1364 & 0.2717	& 0.2188 & 0.1556 & 0.3738 & 0.5102	& \textbf{0.5871} \\
    \hline
    \multicolumn{10}{c}{1: \textit{MaD}\ \ 2: \textit{BasketballPass}\ \ 3: \textit{RaceHorses}\ \ 4: \textit{Vidyo1}\ \ 5: \textit{Vidyo3}\ \ 6: \textit{Vidyo4}\ \ 7: \textit{Kimono}\ \ 8: \textit{TunnelFlag}\ \ 9: \textit{BarScene}\ \ 10: \textit{PeopleOnStreet}}

    \end{tabular}%
  \label{results}%
  \vspace{-2em}
\end{table*}%

\subsection{Training procedure}\label{training}

Since all the four components of our QG-ConvLSTM approach are deep networks, they can be jointly trained in an end-to-end manner. However, the gates generator with an 1D-LSTM structure aims at providing the weights to gate the cells in ConvLSTM, while the other three parts are used to enhance quality. Hence, we first separately pre-train the gates generator, and then all four networks are jointly trained. 

\textbf{Pre-training the gates generator.} As aforementioned, the generated $G_n$ in \eqref{fc} is utilized to weight the ``input'' gate in our quality-gated cell. The frames with high quality are expected to have large $G_n$ to update its advantageous information to $C_n$; contrarily, low quality frames should be with small values of $G_n$. We follow \cite{Yang2018Multi} to account PQFs as high quality frames, and non-PQFs as low quality frames. PQF denotes the frame with peak quality, i.e., the PSNR is higher than the previous and subsequent neighboring frames. Here, let $l_n\in\{0,1\}$ indicate whether the $n$-frame is PQF ($l_n=1$) or not ($l_n=0$). As such, given the raw and compressed videos, the labels $\{l_n\}^{N}_{n=1}$ can be obtained. Then, considering that the number of PQFs is much less than non-PQFs, we apply the weighted sigmoid cross entropy loss
\begin{align}
L_G = &-\frac{1}{N}\sum^{N}_{n=1}(l_n\cdot\log\sigma(K\cdot G_n)\cdot P \nonumber \\
&+ (1-l_n)\cdot\log(1-\sigma(K\cdot G_n))) \label{lg}
\end{align}
to pre-train our gates generator, in which $P$ is the weight to balance the various numbers of positive and negative samples. As expected, when minimizing \eqref{lg}, the $G_n$ values of PQFs are learnt to be larger, while those for non-PQFs are trained to be smaller. Note that, in \eqref{lg}, $\sigma(K\cdot G_n)$ is pushed to approach 0 or 1, and $K>1$ is a scale hyper-parameter, making $f_n = 1 - \sigma(G_n)$ and $i_n = \sigma(G_n)$ not too close to 0 and 1. Otherwise, all previous memory is forgotten or nothing is updated, leading to the ineffectiveness of the temporal structure of our approach.

\textbf{Jointly training the whole network.} After our gates generator converged, the four networks in QG-ConvLSTM are jointly trained. Defining $\{\textbf{X}_n\}^{N}_{n=1}$ as the raw video frames, the loss function of QG-ConvLSTM can be written as
\begin{align}
L_Q = \frac{1}{N}\sum^{N}_{n=1}||\textbf{\^{Y}}_n-\textbf{X}_n||_2^2. \label{lqe}
\end{align}
When jointly training, the gates generator is initialized by the pre-trained model, and can automatically learn the weights to gate our quality-gated cell.
\vspace{-.5em}
\section{Experiments}\label{exp}
\vspace{-.5em}
\subsection{Settings}
\vspace{-.5em}
The experimental results are presented to validate the effectiveness of our QG-ConvLSTM approach. As introduced in Section \ref{pre}, the Vid-70 database \cite{Yang2018Multi} is utilized in this paper. We also follow \cite{Yang2018Multi} to use the training set (60 sequences) of Vid-70 to train our QG-ConvLSTM approach, and test QG-ConvLSTM on the test set (10 sequences) of Vid-70. The training and test sequences are compressed by the latest HEVC standard, using HM 16.0 with the Low Delay P (LDP) mode at QP = 37 and 42.
In our QG-ConvLSTM approach, the layer numbers of the spatial and reconstruction networks are both set as 5, i.e., $M=L=5$. The kernel size of $\{W_{sm}\}^M_{m=1}$, $\{W_{rl}\}^{L}_{l=1}$, $W_c$ and $W_o$ are all set to $5\times 5$, and the channel numbers of all convolutional layers are 24. Then, when obtaining $\{\textbf{Q}_n\}^{N}_{n=1}$, we set $T=4$ in \eqref{qn}. The number of hidden units in LSTM of the gate generator is 256. During training, the Adam algorithm is applied with initial learning rate of $10^{-4}$. The scale parameter in \eqref{lg} is set to 10.

\subsection{Performance of our QG-ConvLSTM approach}

Now, we evaluate the quality enhancement performance of our QG-ConvLSTM approach in terms of $\Delta$PSNR, which measures the PSNR improvement after enhancing the compressed video. Our performance is compared with the state-of-the-art image enhancement methods\footnote{For fair comparison, we re-trained AR-CNN, DnCNN and Li \textit{et al.} on HEVC compressed samples.
}, including AR-CNN \cite{dong2015compression}, Li \textit{et al.} \cite{li2017efficient} and DnCNN \cite{Zhang2017Beyond}, and video enhancement methods of DCAD \cite{Wang2017A}, QE-CNN \cite{yang2017enhancing} and MF-CNN \cite{Yang2018Multi}.

\begin{figure}[!t]
\centering
\vspace{-2.5em}
\includegraphics[width=1\linewidth]{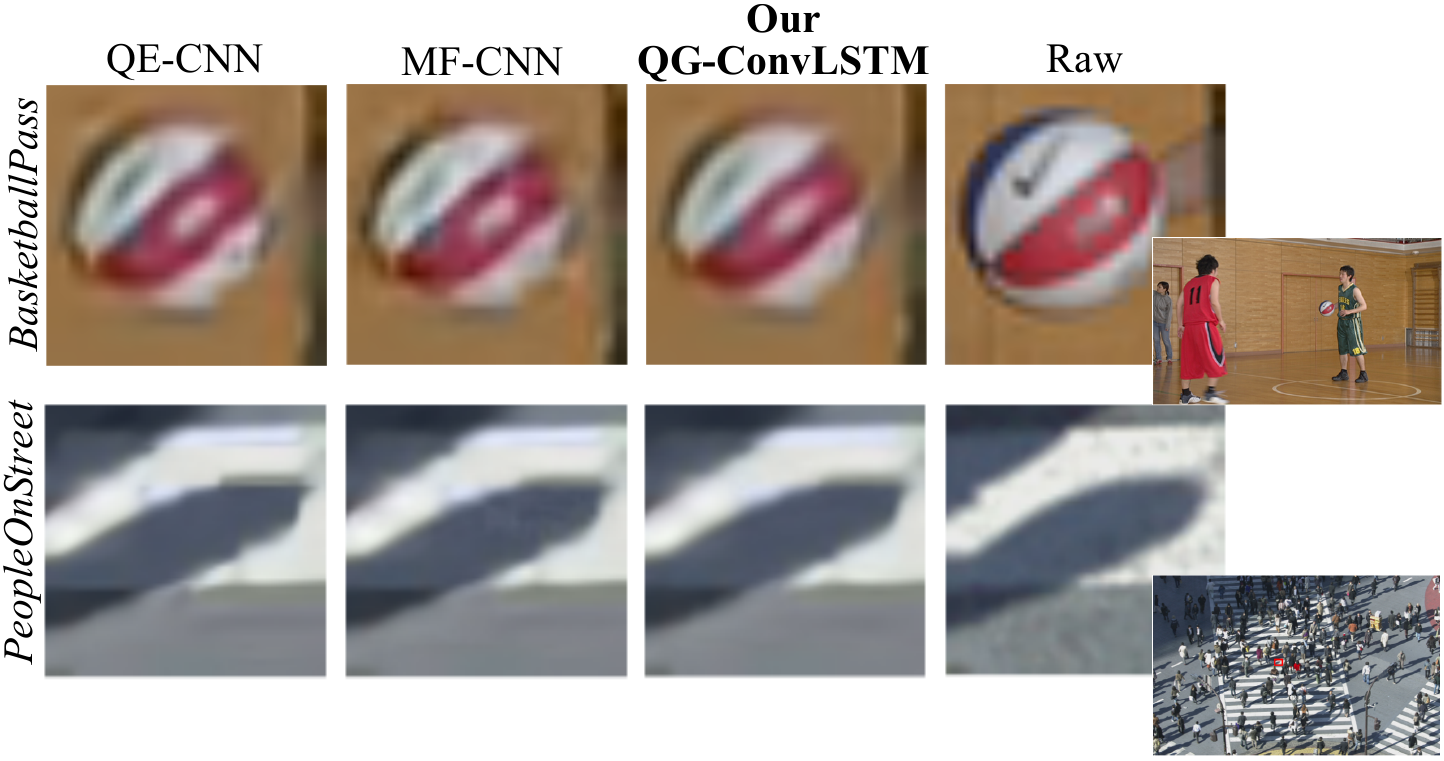}
\vspace{-2.5em}
\caption{Subjective quality of HEVC video}\label{sub}
\vspace{-1.5em}
\end{figure}

Table \ref{results} shows the average $\Delta$PSNR results for all test sequences. As shown in this table, our QG-ConvLSTM approach achieves the best performance on all videos. To be specific, at QP = 42, the highest $\Delta$PSNR reaches 0.9077 dB. The averaged $\Delta$PSNR is 0.6007 dB, significantly higher than the latest video enhancement approach MF-CNN (0.4610 dB), and doubles that of QE-CNN (0.3070 dB). More $\Delta$PSNR gain can be obtained when comparing with the other four approaches. At QP = 37, the $\Delta$PSNR of our QG-ConvLSTM approach (0.5871 dB) is also obviously higher than the second and third best approaches, i.e., MF-CNN (0.5102 dB) and QE-CNN (0.3738 dB), respectively. Besides, the parameter number of our QG-ConvLSTM is only 646,907, which is about 36\% of MF-CNN (1,787,547) and QE-CNN (1,796,514). Accordingly, our QG-ConvLSTM approach outperforms all compared methods on enhancing the quality of compressed video with smaller number of parameters than the latest video enhancement approaches. These validate both the effectiveness and the efficiency of our approach.
Moreover, Fig. \ref{sub} shows the subjective results on videos \textit{BasketballPass} at QP = 42 and \textit{PeopleOnStreet} at QP = 37. It can be seen that our QG-ConvLSTM approach is able to effectively reduce artifacts and achieve the best subjective performance.

\vspace{-.5em}
\subsection{Ablation study}\label{abl}
\begin{figure}[!t]
\centering
\vspace{-2.5em}
\includegraphics[width=.95\linewidth]{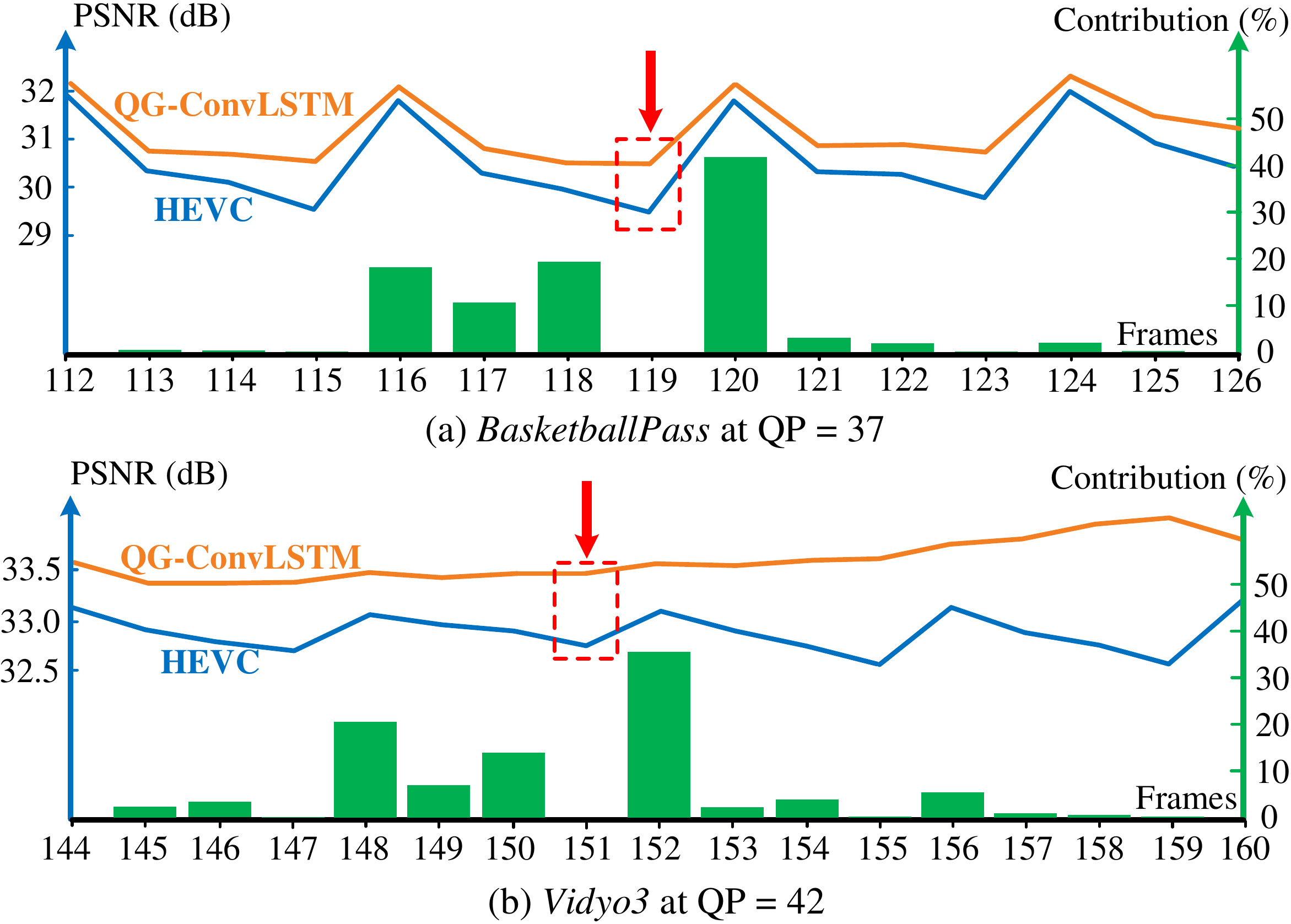}
\vspace{-1em}
\caption{Enhancement performance and the contribution of each frame to enhancing the frames highlighted by red arrays}\label{contr}
\vspace{-1em}
\end{figure}

\vspace{-.5em}
We further analyze the effectiveness of the proposed quality-gated cell in our QG-ConvLSTM approach.
In this paper, the contribution of the $(n$-$k)$-th frame on enhancing the $n$-th frame is calculated by $\frac{1}{2}\cdot i_{n-k}\cdot\prod^{n}_{j=n-k+1} f_j$, and that for frame $(n$+$k)$  is $\frac{1}{2}\cdot i_{n+k}\cdot\prod^{n+k-1}_{j=n} f_j$.
These indicate the proportion of the $(n\pm k)$-th frame information in $C_n$ for enhancing the $n$-th frame. Fig. \ref{contr} presents the contribution of other frames for frame 119 of \textit{BasketballPass} at QP = 37 and frame 151 of \textit{Vidyo3} at QP = 42. It can be seen that our quality-gated cell enlarges the contribution of the frame with high quality and low distance, and reduces the contribution of the low quality frame which is far away. More importantly, the proposed quality-gated cell also learns the trade-off between inter-frame correlation and compressed quality. For instance, in Fig. \ref{contr}-(b), although frame 156 has longer distance with frame 151 than frame 153, because of its higher quality, its contribution is slightly higher than frame 153. As a result, the multi-frame information can be appropriately used for quality enhancement of compressed video.

Besides, we compare the performance with the original ConvLSTM cell \cite{xingjian2015convolutional}. Our experimental results show that utilizing the original ConvLSTM cell instead of our quality-gated cell degrades the average $\Delta$PSNR to 0.5467 dB and 0.5272 dB for QP = 37 and 42, respectively.
These prove the effectiveness of the proposed quality-gated cell in our QG-ConvLSTM approach.

%
\vspace{-1em}
\section{Conclusion}
\vspace{-.5em}
This paper has proposed the QG-ConvLSTM approach for enhancing the quality of compressed video. Our QG-ConvLSTM approach applies the bi-directional ConvLSTM structure to fully exploit the advantageous information for quality enhancement in the video. More importantly, we proposed learning the ``forget'' and ``input'' gates in ConvLSTM from quality features, thus the weights to forget the previous memory and to update the current information are guided by compressed quality. As such, frames with various quality contribute differently to helping other frames to enhance quality. Consequently, the multi-frame information in the video can be adequately and reasonably used. Finally, the experiments validated that our QG-ConvLSTM outperforms other state-of-the-art quality enhancement approaches.

\bibliographystyle{IEEEbib}
\bibliography{icme2019template}

\end{document}